# A Physiology-Driven Computational Model for Post-Cardiac Arrest Outcome Prediction

Han B. Kim[1†], Hieu T. Nguyen[1†], Qingchu Jin[1†], Sharmila Tamby[2†], Tatiana Gelaf Romer[1†], Eric Sung[1†], Ran Liu[1], Joseph L. Greenstein[1], Jose I. Suarez[3,4,5], Christian Storm[6], Raimond L. Winslow[1], Robert D. Stevens[3,4,5]

[1] Department of Biomedical Engineering, Whiting School of Engineering, Johns Hopkins University, Baltimore, Maryland, USA.

[2] Department of Computer Science, Whiting School of Engineering, Johns Hopkins University, Baltimore, Maryland, USA.

[3] Department of Anesthesiology and Critical Care Medicine, Johns Hopkins University School of Medicine, Baltimore, Maryland, USA.

[4] Department of Neurology, Johns Hopkins University School of Medicine, Baltimore, Maryland, USA.

[5] Department of Neurosurgery, Johns Hopkins University School of Medicine, Baltimore, Maryland, USA.

[6] Department of Nephrology and Intensive Care Medicine, Charité-Universitätsmedizin, Berlin, Germany.

[†]These authors contributed equally to this work as co-first authors.

*Corresponding author: Robert D Stevens, MD, Johns Hopkins University School of Medicine, email: rstevens@jhmi.edu

**One Sentence Summary**

Integrating early physiological time series features extracted from a very large multicenter database of intensive care unit patients, we developed and validated machine learning models which accurately predicted mortality and neurological outcome of patients resuscitated from cardiac arrest.

**Abstract**

Patients resuscitated from cardiac arrest (CA) face a high risk of neurological disability and death, however pragmatic methods are lacking for accurate and reliable prognostication. The aim of this study was to build computational models to predict post-CA outcome by leveraging high-dimensional patient data available early after admission to the intensive care unit (ICU). We hypothesized that model performance could be enhanced by integrating physiological time series (PTS) data and by training machine learning (ML) classifiers. We compared three models integrating features extracted from the electronic health records (EHR) alone, features derived from PTS collected in the first 24hrs after ICU admission ($PTS_{24}$), and models integrating $PTS_{24}$ and EHR. Outcomes of interest were survival and neurological outcome at ICU discharge. Combined EHR-$PTS_{24}$ models had higher discrimination (area under the receiver operating characteristic curve [AUC]) than models which used either EHR or $PTS_{24}$ alone, for the prediction of survival (AUC 0.85, 0.80 and 0.68 respectively) and neurological outcome (0.87, 0.83 and 0.78). The best ML classifier achieved higher discrimination than the reference logistic regression model (APACHE III) for survival (AUC 0.85 vs 0.70) and neurological outcome prediction (AUC 0.87 vs 0.75). Feature analysis revealed previously unknown factors to be associated with post-CA recovery. Results attest to the effectiveness of ML models for post-CA predictive modeling and suggest that PTS recorded in very early phase after resuscitation encode short-term outcome probabilities.

## Introduction

Cardiac arrest (CA) is an abrupt cessation of myocardial function which affects more than half a million people in the United States annually. An estimated 80% of patients are unconscious after resuscitation from CA, and these patients may experience a wide range of outcome trajectories, from complete recovery to death or severe neurologic disability *(1)*. A major challenge in post-CA care is to accurately predict outcome, especially in the early phase after CA when patients are treated in the intensive care unit (ICU). Current guidelines recommend that prognostication should integrate physical examination findings with neurophysiologic tests and other tests (multi-modality prognostication), and should be delayed until at least 72 hours after CA *(2)*. Multi-modality prognostication can however be a challenge to implement, and the predictive performance of its different elements, while studied individually, are unknown in aggregate. One of the most widely used statistical models to assess patients admitted to the ICU is the Acute Physiologic Assessment and Chronic Health Evaluation (APACHE) *(3)*. The APACHE score is a logistic regression model which combines clinical and laboratory variables present on ICU admission to provide an estimate of severity of illness and short-term mortality. The predictive performance of APACHE is however limited due to oversimplified regression modeling and because it may not account for disease-specific covariates *(4)*. Here, we propose a novel approach for post-CA prediction, based on two hypotheses. First, that integrating widely available continuous physiological time series (PTS) data will make prognostic models more accurate than models without PTS. And second, that machine learning (ML) classifiers will perform better than the APACHE III logistic regression model. Hypotheses were tested using data from the multicenter eICU-Clinical Research Database (hereafter referred to as eICU), which consists of time-series, numerical, and categorical data on >200,000 ICU admissions in 208 hospitals across America *(5)*.

## Results

Data were extracted on 2,216 CA patients meeting inclusion criteria (Fig. 1). A demographic summary of the patient population is provided in Table 1 and the distribution of patients according to outcome is in Table 2.

Discrimination (AUCs using the nested resampling method [see Methods]) of the different ML models implemented to predict neurological outcome is shown in Fig 2 and Table S1. APACHE III was computed with the exact features used in clinical practice and modeled in two ways: logistic regression (APACHE Generalized Linear Model [GLM]), and random forest (APACHE RF). While the APACHE GLM model is used in clinical practice; we implemented APACHE RF to investigate whether the same APACHE features could achieve a higher AUC when used to train a RF classifier. The RF APACHE outperformed the GLM by a small margin of 2.1%. The remaining five models use the complete set of features, both EHR and $PTS_{24}$, as inputs. The performance of RF, GLM with elastic net regularization, gradient boosting (XGBoost), and Neural Network (NN), and an Ensemble model are evaluated. The Ensemble model is an average of the four-best performing first-level models (XGBoost, GLM Elastic Net, RF, and NN). Each model outperformed both versions of APACHE by over 8%. Most notably, the Ensemble model, the best performing model, reached a mean AUC of $0.87 \pm 0.01$, outperforming APACHE by 12.3% and as can be seen in table 3, the difference in results were statistically significant based on the Wilcoxon ranked sum test.

Discrimination of the different ML models to predict post-CA survival is in Fig. 3 and Table S2. AUC of the APACHE RF was 2.7% higher than the APACHE GLM. Models utilizing EHR and $PTS_{24}$ had a mean AUC ranging from 0.81 to 0.85. The Ensemble model achieved the highest performance, with a mean AUC of $0.85 \pm 0.01$, surpassing the APACHE GLM baseline by 14.5%. Despite similar model performances, it is clear from the feature rankings (figs. 4 and S4) that individual outcome prediction models ranked features differently. Table 3 and Fig. 4 summarizes the best performing model (BPM) for both neurological outcome and survival prediction and Table S1 and S2 shows statistically significant differences in discrimination which was based on adjacent Wilcoxon ranked sum tests. Of note, the NN model underperforms when compared to other ML algorithms (Figs. 2, 3 and Tables S1, S2).

***Feature Subset Performance Comparison***

Performance of the models for prediction of post-CA neurological outcome and survival using different feature subsets is shown in Fig. 4A and in Tables 4 and S3. The AUCs for the three sets of features (EHR, PTS$_{24}$, and combined HER+PTS$_{24}$) are shown in Fig. 4A. For neurological outcome prediction, the APACHE model had the lowest performance, with an AUC of 0.75, sensitivity of 0.77, and specificity of 0.63. The PTS$_{24}$-only model showed slightly improved performance, reaching an AUC of 0.78, a sensitivity of 0.66, and a specificity of 0.74. The EHR-only model (inclusive of the APACHE features) reached an AUC of 0.83, a sensitivity of 0.77, and specificity of 0.77. Finally, the BPM combining EHR and PTS$_{24}$ features had an AUC of 0.87, sensitivity of 0.78, and a specificity of 0.88. The addition of PTS$_{24}$ features to EHR increased the AUC by 3.7%, enhancing model performance by a margin that was statistically significant (Table S3). The tabulated performance metrics and results for statistical significance are in Table 4. All three feature subsets outperformed APACHE in terms of AUC and specificity, however, the sensitivity remained similar or worse.

Similar performance metrics were found when comparing feature subsets for the prediction of post-CA survival (Figure 4B and Tables 5 and S3). PTS$_{24}$ alone achieved an AUC of 0.68 which was lower than the reference APACHE model. However, the addition of PTS$_{24}$ features to EHR increased the AUC by 4.5% in the combined EHR and PTS$_{24}$ model, which was a larger contribution to model performance than in the neurological outcome prediction model. The impact of PTS$_{24}$ on model performance for both principal outcomes supports our hypothesis that physiological features recorded in the first day after ICU admission are associated with post-CA recovery.

*Model Interpretability*

To identify and rank features with the highest predictive value in our ensemble model we used random forests to compute each feature's minimum and maximum tree depth. We defined relative importance (RI) as a range from 0 to 1 with 1 being the most important feature; features with higher RI have lower minimum tree depth and are more predictive of outcome (for details see Methods section). Each feature was categorized to determine what variable types were most predictive of neurological outcome (Figs. 5 and S4). Several of the highest-ranked features belong to the APACHE variable category and lab results category including initial GCS scores, highest temperature, and lactate levels. Heart rate and oxygen saturation PTS features were more predictive compared to other PTS features derived from diastolic blood pressure, systolic blood pressure and respiratory rate. We performed the same feature ranking for prediction of survival (Fig. S4). As with neurological outcome prediction, GCS total and subscores ranked as most important features but other lab results such as glucose, albumin, monocytes, MCV, sodium, and protein had greater RI. Additionally, patient information such as age, BMI, weight, height, time from admission, ranked much higher than in the neurological outcome model feature ranking.

To better understand the impact of top-ranked feature, we assessed the correlation of each top-ranked feature with neurological outcome by performing a univariable analysis of the top ranked 20 features using the coefficients from a generalized linear model (Table 6). Table 6 contain some statistical features engineered from $PTS_{24}$ with the help of a statistical feature extraction package HCTSA *(6)* which generated hundreds of statistical features per $PTS_{24}$ signal; the latter were pruned by selecting the most impactful feature per $PTS_{24}$ signal based on random forest feature ranking. Features which were positively correlated with favorable neurological outcome included: initial total GCS at admission, minimum motor, eye, and verbal GCS subscores on day 1, maximum temperature on day 1, heart rate variability surrogate (mean of 20 time-binned means of 24 hrs.), and dexmedetomidine drug use (Table 6). Mean and max lactate level within day 1, and SOFA score were negatively correlated with favorable neurological outcome, consistent with what has been previously reported in the literature *(7)*.

In addition to the analysis RI across features, we used a recurrent neural net (RNN, see Methods) to explore the time dependence of PTS data during the first 24 hrs. after admission to the ICU (Fig. 6). An attention layer in the RNN was designed to show the contribution of PTS data each hour after admission. As shown in Fig. 6, the first four hours of PTS data contributed most to prediction.

In order to leverage PTS data from a much larger sample of eICU patients outside the ones with CA selected for the primary analysis, we implemented a transfer learning neural network model. We pretrained the convolutional neural network (CNN) model on the survival task using 140,200 non-CA patients and then further trained the model on our target tasks: neurological outcome or survival of CA patients. The pretrained model on survival prediction of survival in non-CA patients achieved a highest test-set AUC of 0.9 utilizing both EHR and $PTS_{24}$ data, and attained 0.87 and 0.85 test-set AUC when trained on only EHR data and $PTS_{24}$ data, respectively (Table S5). The pretrained model was then re-trained (transferred) on the CA-patient population, which improved the performance by 2.5-3% in test-set AUC in models using PTS$_{24}$ data alone (Fig. S3). The transferred model was then incorporated in the final NN model shown in Fig. 2 and Table S1 (for neurological outcome prediction) and Fig. 3 and Table S2 (for post-CA survival).

**Discussion**

This work supports our two main predictions, first that physiological features in the first 24 hrs after ICU admission increase the accuracy of outcome prediction in patients following CA; and second, that prediction accuracy is enhanced with models that use ML classifiers compared to widely used logistic regression models. When compared to the EHR model, PTS$_{24}$ increased discrimination of the best performing neurological outcome prediction model by 3.7%, while discrimination of the survival prediction model increased by 4.5%. Collectively, these findings indicate that time-series feature engineering is a promising approach to decode high-frequency physiological data, providing unique insight on the condition of patients after CA which are valuable in predicting clinically meaningful outcomes. Moreover, our analyses were performed using data which are routinely acquired for clinical purposes in ICUs around the world, suggesting opportunities for validation and implementation on a larger scale.

We found that the highest performance was achieved with an ensemble model combining features from both EHR and PTS$_{24}$ and averaging the four best performing first-level models (XGBoost, GLM Elastic Net, RF, and NN). This suggests that each of these models integrates slightly different, complementary information which combined produces greater discrimination than each of the first level models alone. For neurological outcome prediction the ensemble method outperformed the other models by as much as 3.5%. While the ensemble method has a similar sensitivity to APACHE GLM (sensitivity = 0.78), it improved the specificity by 25% and the AUC by 12.7%. An almost identical pattern can be seen with the ensemble model for survival prediction. The latter achieved a 15% increase in AUC and 19% increase in specificity when compared to the reference APACHE.

Since the data for our CA patients originate from over 200 hospitals, each with idiosyncrasies and variability in clinical practices, we believe that our models for neurological outcome and survival predictions are highly generalizable to post-CA patients within any US hospitals. There is considerable variability between medical institutions (and even between ICUs within the same institutions) in the way intensive care medicine is structured and delivered, and this variability is also seen in the acute care of patients resuscitated from CA. By creating our post-CA outcome models based on the variable CA patients from many institutions, our model can be generalized and applied to a large variety of CA patients without drastic performance loss or the need to retrain the model to work for a single institution.

In our exploration of the feature space, we identified several prognostic variables which are well-known, and others which have not been previously reported. A number of factors have been associated with post-CA neurological outcomes *(8–11)*. We found that the level of consciousness as measured by the GCS (total and subscores) appeared several times in the top ranked features (Table 6). In addition, the highest ranked features included surrogate measures of the statistical variations of heart rate variability and fluctuation. We found that a surrogate feature for heart rate variability was positively correlated with favorable neurological outcome. Specifically, this surrogate feature is a measure of 20 bins of heart rate signal recorded over 24 hours, capturing fluctuations in heart rate that occur every 1.2 hours. Results indicated that patients with a greater surrogate heart rate variability statistic were more likely to have a favorable neurological outcome. Heart rate variability has been linked to cardiovascular disease risk and outcomes but was not previously known to be associated with post-CA neurological outcome *(12, 13)*. Maximum temperature recorded in the first 24 hours and dexmedetomidine infusion use were also unexpectedly correlated to favorable outcomes. Dexmedetomidine is commonly used as a sedative infusion for patients in the ICU. The correlation observed here suggests a protective effect of dexmedetomidine which might need to be further investigated in this population. Additionally, we found that neither blood pressure nor pulse oximetry time series correlated strongly with outcome. This might reflect the fact that hemodynamic and respiratory variables are tightly controlled in the ICU setting, lessening their inferential physiological value; a more meaningful approach might have been to study the blood pressure time series in the context of vasoactive infusions and of mechanical ventilation settings, data which were unfortunately not readily available in eICU.

This research also provided insight on the time-dependence of PTS data for outcome prediction. Physiologic features recorded within the first two hours following ICU admission seem most influential in predicting neurological outcome. Prior studies using techniques such as continuous EEG or Bispectral index have suggested a relationship between very early neurophysiological signals and neurological outcome *(14, 15)*.

To our knowledge, the results we present using transfer learning are the first time this technique has been successfully implemented on medical time series data. The advantages of utilizing transfer learning with CNN models as been demonstrated in other settings *(16, 17)*. The gain in discrimination (3% AUC increase) for the models trained on the $PTS_{24}$ data suggests that transfer learning could be helpful in extracting useful latent information from unstructured medical time series data, particularly applied to subpopulations in which data is limited. Interestingly, transfer learning had a negligible effect on performance of the models trained on the EHR data alone or on the $PTS_{24}$+EHR data. We postulate that it could be due to two factors. The first is that the fully connected layered architecture in the neural net for the structured EHR data was a simple task such that even without transfer learning, the neural net could still achieve optimal performance. The real value of transfer learning is noted when the data are unstructured like PTS or images, which involved fine-tuning convolutional layers. The second factor as to why transfer learning did not improve model performance for the EHR+ $PTS_{24}$ data could be because of the way EHR and PTS are coupled in our neural net architecture. This neural net collects the output (predicted probability of having the outcome) from the EHR sub-network and from the PTS sub-network and combines the two outputs into a final predicted probability output via a fully connected layer and a sigmoid function. This architectural design might not be the optimal way to combine EHR and PTS data. Nonetheless, our results still show that transfer learning did help increase performance when the input data is PTS alone, a finding which has not been reported before in studies dealing with medical time-series.

### *Limitations*

Several limitations must be noted in this work. This is a retrospective analysis suggesting potential errors due to confounding and bias. Specifically, no information was available about withdrawal of life sustaining therapy (WTST); yet it is known that WLST occurs in a significant proportion of CA patients and can express self-fulfilling prophecies *(18)*. Furthermore, the preprocessing steps undertaken with the raw eICU data may not reflect the reality of a patient's dynamic health state in the ICU. Missing data were imputed using a variety of methods (described in the Methods section) but include the possibility of error.

Another limitation is that validated post-CA outcome measures such as the cerebral performance category (CPC) score were not available in the eICU database. This is a significant limitation because the CPC score is the outcome measure most widely used in this population. We used a surrogate outcome parameter based on the motor subscore of the GCS to estimate neurological function; however, the mGCS does not capture the range of functional states that is encompassed in scores like the CPC *(8–10)*. Additionally, our outcomes were assessed at discharge from the ICU, an early time point beyond which significant recovery of function may still occur. Last, although the HCTSA package was implemented to exhaustively extract time series features to explore the full potential of PTS data, many of the derived statistics were challenging to interpret in terms of clinical significance. However, we found this approach to be valuable, especially when compared to artificial neural networks which generally lack interpretability *(19)*.

### *Future work*

All our analyses were performed retrospectively and without any of the time constraints that are relevant to decision-making in the ICU. An important goal will be to validate our results on prospectively collected data to establish the efficacy, generalizability and practicability of this approach in a real-world (and real-time) setting. In addition, work is needed to understand the relevance of such models to long-term outcomes, and to determine if comparable predictions are possible earlier in the ICU admission, other data extraction and analysis time frames will be considered. Preliminary results obtained with data from the first 2, 6, 12, and 18-hour time windows after ICU admission suggest that it may be possible to attain similar predictive power as early as 2 hours after admission with only a slight decrease in predictive performance. Moreover, we plan to conduct further investigations into treatment responsiveness for post-CA patients. Targeted Temperature Management (TTM) is a treatment that is commonly implemented, in which mild hypothermia is induced to limit the extent of brain damage in post-CA patients. While TTM has been shown to significantly improve both survival and neurological outcomes *(7, 20)*, it is associated with risks and requires considerable resources and labor to be deployed in the ICU *(21)*. Moreover, it is known that the benefits and risks of TTM vary considerably across CA populations. We plan to identify subgroups of CA patients who will respond favorably to TTM to help individualize therapy.

In conclusion, we demonstrate that ML models can achieve state-of-the-art prediction of early neurological outcome and survival of patients resuscitated from CA. Additionally, our work confirms that time-series physiological features significantly enhance the performance of clinical prediction models.

## Materials and Methods

We evaluated seven ML algorithms (generalized linear model [GLM], random forest [RF], gradient boost [XGBoost], and neural network [NN]) trained on different feature subsets (APACHE III features, selected EHR features, PTS features, and combined EHR and PTS features), to predict one of two clinical outcomes (neurological outcome and survival, both assessed at discharge from the ICU). APACHE III features are a subset of the EHR feature space and do not contain any PTS features. To determine the performance of each model, the sensitivity, specificity, and AUC were computed across five inner and five outer validation loops (Fig. 7 and described below). To obtain an accurate comparison with the reference APACHE III, all models were generated using exclusively features extracted from the first 24 hours following ICU admission.

### *Feature and label space definition*

The Philips eICU Research Institute (eRI) maintains an open source database (eICU version 2.0) containing over 200,859 unique patient encounters in 208 hospitals that use tele-ICU software across the US *(5)*. The detailed demographic information of the eICU database is shown in Table

3. This data consists of patients who were admitted to ICUs in 2014 and 2015. The database is released under the Health Insurance Portability and Accountability Act (HIPAA) safe harbor provision. The re-identification risk was certified as meeting safe harbor standards by Privacert (Cambridge, MA) (HIPAA Certification no. 1031219-2).

The sample selection process is illustrated in Fig. 1. Patients were included in the analysis if they were admitted to the ICU after CA, remained in the ICU for >24 hrs., were endotracheally intubated, and if the principal outcome measure was recorded ≤24 hrs. before discharge from the ICU. We included only patients who were intubated and mechanically ventilated as they represent a subset with higher severity of illness in whom prognostication is most relevant *(22)*.

The two outcomes of interest were survival and neurological function at the time of discharge (or ≤ 24 hrs before discharge) from the ICU (longer term outcomes are not recorded in eICU). Survival status was available for all patients. The neurological outcome indicator most widely used in the CA population is the Cerebral Performance Category (CPC) score *(23)*, however this was not available in the eICU database. Neurological outcome was therefore defined based on the motor subscore of the Glasgow Coma Score (mGCS) which was dichotomized as follows: mGCS of 6 (favorable outcome), mGCS ≤5 (unfavorable outcome). Ultimately, a patient sample of 2,216 was defined. For neural networks, raw PTS data is used directly, so patients without adequate PTS data were removed; 1,917 patients met this criterion and their data was used in the neural network analysis.

### *Feature Selection*

A selection was made of EHR features relevant to CA in the ICU setting such as co-morbidities, clinical indicators, laboratory results, and fluid intake/output. To identify new features not previously known to be associated with CA, we extracted nearly all variables available in eICU. Next, to determine the impact of PTS features on prediction when combined with EHR features, a statistical feature extraction was performed using a preexisting open-source time series data processing package available in MATLAB (Highly comparative time-series analysis [HTCSA] *(6)*. This package extracted over 3000 different statistically derived features from the PTS data including distribution, correlation, and trend properties over the entire dataset and various time windows *(24–27)*.

### EHR Preprocessing

EHR variables for which > 40% of data were missing were excluded from the analysis. For the remaining variables, random forest unsupervised imputation was used to fill missing values. The method, fully described in *(28)*, is relatively fast and takes the nonlinearity and interaction among variables into account. The method was implemented with five iterations using the R package 'RandomForestSRC.'

### PTS data description and preprocessing

PTS data in the eICU database is recorded as a windowed median every 5 minutes. We extracted heart rate (HR), systolic and diastolic blood pressure, respiratory rate (RR), and O2 saturation by pulse oximeter (SpO2). PTS data was extracted only for the first 24 hours (PTS$_{24}$). Because PTS data can be highly irregular, pre-processing steps were needed to standardize and impute the data. Preprocessing can be separated into two parts: clinically implausible data (outlier) detection and missing value imputation. We used clinician-set boundaries (Table S4) and statistical methods to detect outliers. Missing PTS data was imputed using EHR data when possible, then linearly interpolated. Additional details are in the supplements and Fig. S1.

### Feature Ranking

Feature ranking and selection was performed to increase interpretability, reduce model complexity, and avoid overfitting. The main goal was to extract the highest ranked features for use in our final model. We implemented RF, which establishes hierarchical importance for each feature estimated by the frequency and placement (tree depth) of each feature in each decision tree *(29, 30)*. Features with less tree depth are more important than those with greater tree depth. Then, these ranked features were correlated with the outcome to further increase interpretability. This ranking was then normalized to a range of 0 to 1.

Time-dependent importance of PTS data was studied with a recurrent neural network (RNN). We implemented six gated recurrent units (GRUs) in parallel. Each gated recurrent unit is connected to an attention layer (α), shown in Fig. S2E. The length of the attention layer is designed as 24 to represent the 24 hours. The weights in the attention layer were normalized and the summation across all time steps is one. The weights in the attention layer are designed to show the contribution of PTS data at different time steps on the prediction. The rationale is that, through training, the RNN optimizes the attention layer to maximize the performance such that the model spontaneously assigns higher weights to time where there is a larger contribution of prediction. This technique has been used successfully for other classification and prediction tasks *(31, 32)*. An average weights vector was obtained from six paralleled attention layers to assign time-dependent importance. We calculated and averaged the weights of all patients to plot Fig. 6.

### *Supervised Learning Pipeline*

Fig. 7 illustrates training and evaluating of our models using a nested cross-validation method containing two cross-validation loops. The inner loop is for hyperparameter tuning, while the outer, 5-fold x 5 times loop is used to estimate the generalized performance and to compare performances of different models. We used this nested cross-validation method instead of the traditional non-nested cross-validation method to avoid overestimation of the true performance. With the traditional k-fold cross-validation, the same data is often used for both for hyperparameter tuning and for estimating the generalization error, which can lead to information leakage and overfitting. Studies have demonstrated that the non-nested, cross-validated error estimate for the classifier with the optimal parameters is a substantially biased estimate of the true error that the classifier would incur on another, independent dataset *(33–35)*. The nested cross-validation resampling strategy has been shown to be an optimal estimator of the true error *(34, 36)*.

*Machine Learning Algorithms*

Several ML methods were used to train the prediction models. Four ML models were implemented using the 'caret' package in R. First, 'glmnet' was selected to implement a generalized linear model that uses a penalized maximum likelihood method and elastic net penalty for the regularization path *(37)*. Next, 'ranger' was selected for random forest due to its fast implementation and successful past performance with high dimensional data *(38)*. We also used the optimized extreme gradient boosting ML library, XGBoost, because it has been successfully used for classification tasks and implements a regularized model to avoid overfitting *(39)*.

We tested several well-known neural network (NN) architectures including fully connected, convolutional, and recurrent layers. Because of the difference of intrinsic data properties, fully connected layers were used for structured EHR data and HCTSA-derived PTS features. Convolutional layers (CNNs) and Recurrent layers (RNNs) were used for unstructured PTS data. Two widely used RNN structures were tested: long-short-term memory (LSTM) and Gated recurrent unit (GRU) *(40)*. All deep learning code was implemented using the Python 3.7, 'pytorch 0.4.1.post2' package. Model training was performed through the Maryland Advanced Research Computing Center cluster with 2 Tesla P100 GPUs. We implemented an RNN with an attention layer allowing us to evaluate the relative importance of PTS data at each hour. Detailed description of RNN is in the Supplement.

*Model Optimization*

For each model, hyperparameters were tuned in the inner loop of nested cross-validation (10-fold x 3 times, see Fig. 7). Hyperparameters for most models were tuned using the grid-search method with default hyperparameter space in the 'caret' package. In addition, for the best performing first-level models (such as XGBoost, GLM-elastic net, RF, and NN), Bayesian (model-based) optimization was implemented using the 'mlrMBO' package in R and bayes_opt package in Python *(41, 42)*. Bayesian optimization has been established as an efficient and automatic hyperparameter tuning method that produces state-of-the-art performance *(42, 43)*.

To mitigate the overfitting inherent in the use of NNs, transfer learning was implemented. Transfer learning utilizes knowledge of previously trained models from a different feature space with a different distribution and transfers this knowledge to the current model as initial weights *(44)*. This method has seen wide success in image classification tasks *(45)* as well as time series-related tasks *(46)* by first training on another pre-task with numerous data to produce initial weights *(47–49)*. The weights from the pre-trained model are then applied to the new classification task, in a related target domain. Transfer learning can improve generalizability across different distributions and improve models that have a small sample space. In this study, the pre-training is conducted on the entire available patient population with available features after preprocessing (~140,000 patients), with survival (related target domain) as the label. The weights from the pre-trained neural network are then transferred to trained on our target population of 1,917 CA patients, to predict both survival and neurological outcome.

***Model Stacking***

To further boost predictive performance, we combined the predictions of different ML models in a single Ensemble model. Model stacking is an efficient ensemble method in which predictions generated by different learning algorithms are used as inputs in a second-level learning algorithm *(50, 51)*. For the second-level model, different algorithms were applied, from a simple unweighted average of predicted probabilities from each first-level model, to GLM, RF, and XGBoost. We found that all these second-level approaches produced similar results for our dataset, thus the simplest averaging predicted probability method was chosen for its minimum computational expense. In our study, the final Ensemble model is an average of four of our best performing first-level models: XGBoost, GLM Elastic Net, RF, and NN.

**Supplementary Materials**

Materials and Methods

Supplementary Tables and Figures

Table S1. Statistical comparison of the discrimination of the seven machine learning algorithms used to predict post-CA neurological outcome.

Table S2. Statistical comparison of the performance of the seven machine learning algorithms used to predict post-CA survival.

Table S3. Statistical comparison of model performance using different feature sets.

Table S4. Clinically determined lower and upper bounds for physiological time series.

Table S5. Transfer learning pre-trained model results (survival prediction).

Fig. S1. Physiological time series denoising and imputation.

Fig. S2. Architecture of neural networks used in this study.

Fig. S3. Discrimination of transfer learning models.

Fig. S4. Relative importance of feature categories for survival prediction.

**Acknowledgments**

We thank Stephen Granite for his help accessing the eICU database, and Sridevi Sarma for her constructive feedback at the group presentations. **Funding:** None. **Author contributions:** RS, RW, CS and JS formulated the research questions and designed the study. HK, HN, QJ, ST, TR, and ES queried data, constructed the methods, built the models, performed statistical analysis, and wrote the manuscript. RL, JG, and RW provided engineering input and edited the manuscript. RS, JS, and CS provided clinical input and edited the manuscript. **Competing interests:** The authors declare that they have no competing interests. **Data and materials availability**: Data analyzed in this study are publicly available from the eICU database (https://eicu-crd.mit.edu/). To enhance reproducibility, all code used to extract, process, and analyze data will be made available on Github. All external R libraries used for data extraction, processing, or analysis are available via CRAN (https://cran.r-project.org/). Analysis using deep learning utilized Pytorch (https://pytorch.org/).

**Tables and Figures**

**A Physiology-Driven Computational Model for Post-Cardiac Arrest Outcome Prediction**

Han B. Kim, Hieu Nguyen, Qingchu Jin, Sharmila Tamby, Tatiana Gelaf Romer, Eric Sung, Ran Li, Joseph Greenstein, Jose I. Suarez, Christian Storm, Raimond Winslow, Robert D. Stevens

**Table 1. Patient demographic summary**

|  | Favorable Neurological Outcome | Unfavorable Neurological Outcome | P-values |
|---|---|---|---|
| Age (mean [SD]) | 61.93 (15.33) | 62.94 (16.18) | 0.129 |
| BMI (mean [SD]) | 30.21 (9.27) | 33.31 (9.45) | 0.689 |
| Sex, male (n [%]) | 1,681 (61.71) | 1,811 (55) | 8.13e-3 |
| mGCS on admission (mean [SD]) | 4.20 (1.95) | 2.34(1.80) | 2.20e-16 |
| Ethnicity (n [%]) |  |  |  |
| African American | 158 (15.11) | 210 (17.95) | 0.0821 |
| Asian | 10 (0.96) | 16 (1.37) | 0.483 |
| Caucasian | 762 (72.85) | 800 (63.48) | 0.0239 |
| Hispanic | 33 (3.15) | 34 (2.91) | 0.827 |
| Native American | 9 (0.86) | 15 (1.28) | 0.452 |
| Unknown | 74 (7.07) | 95 (8.12) | 0.457 |
| Patients on dialysis | 78 (7.46) | 70 (5.98) | 0.192 |
| Patients on ventilator | 882 (84.32) | 1078 (92.14) | 1.35e-08 |
| Patients with shockable initial rhythm | 169 (16.16) | 86 (7.35) | 6.80e-10 |
| Patients with unshockable initial rhythm | 207 (19.79) | 312 (26.67) | 5.94e-4 |

mGCS: motor Glasgow Coma Scale subscore; BMI: body mass index

| Table 2. Distribution of clinical outcomes | |
|---|---|
| Favorable Neurological Outcome | 1,046 |
| Unfavorable Neurological Outcome | 1,170 |
| Total | 2,216 |
| | |
| Alive | 1,322 |
| Died | 894 |
| Total | 2,216 |

Favorable outcome designates patients whose mGCS was 6; unfavorable neurological outcome designates patients whose mGCS was <6 or who died. All outcomes are at the time of discharge from the ICU

**Table 3. Best performing models to predict neurological outcome and survival**

| Data type | Neurological outcome prediction | | | Survival prediction | | |
|---|---|---|---|---|---|---|
| | APACHE GLM | BPM | Difference | APACHE GLM | BPM | Difference |
| AUC | 0.75 | 0.87 | +0.13 | 0.70 | 0.85 | +0.15 |
| Sensitivity | 0.77 | 0.78 | +0.01 | 0.86 | 0.80 | -0.06 |
| Specificity | 0.63 | 0.88 | +0.25 | 0.56 | 0.75 | +0.19 |

AUC: area under the receiver operating characteristic curve; APACHE: Acute Physiology and Chronic Health Evaluation; GLM: generalized linear model, BPM: best performing model (Ensemble model). See also Fig. 3

**Table 4. Performance metrics of different feature subsets for neurological outcome prediction**

|  | APACHE | EHR | PTS | EHR + PTS |
|---|---|---|---|---|
| AUC | 0.75 | 0.83 | 0.78 | 0.87 |
| Sensitivity | 0.77 | 0.77 | 0.66 | 0.78 |
| Specificity | 0.63 | 0.77 | 0.74 | 0.88 |

EHR: electronic health record, PTS: physiological time series, AUC: Area under the receiver operating curve

**Table 5. Performance metrics for different feature subsets for survival prediction**

|  | APACHE | EHR | PTS | EHR + PTS |
|---|---|---|---|---|
| AUC | 0.70 | 0.80 | 0.68 | 0.85 |
| Sensitivity | 0.74 | 0.76 | 0.78 | 0.80 |
| Specificity | 0.64 | 0.72 | 0.56 | 0.75 |

EHR: electronic health record, PTS: physiological time series, AUC: Area under the receiver operating

**Table 6. Ranking of top 20 features for neurological outcome prediction**

| Ranking | Feature type | Correlation |
|---------|--------------|:-----------:|
| 1 | GCS on admission | + |
| 2 | Minimum mGCS | + |
| 3 | Minimum eGCS | + |
| 4 | Maximum temperature on day 1 | + |
| 5 | SOFA score | - |
| 6 | Minimum verbal GCS on day 1 | + |
| 7 | Mean lactate level | - |
| 8 | HR: Mean of 20 binned means of 24 hrs [surrogate of HR fluctuation] | + |
| 9 | Dexmedetomidine Used | + |
| 10 | Max lactate level on day 1 | - |
| 11 | Mean monocyte level | + |
| 12 | Heart rate POLVAR 5,5 [measure of the probability of obtaining the same consecutive value] | - |
| 13 | Heart rate permutation entropy 4,2 [measure of complexity based on sequences of ordinal patterns dictated by m: the order of entropy, and tau: the time delay for sequence detection] | + |
| 14 | Maximum monocyte count on day 1 | + |
| 15 | Last monocyte count on day 1 | + |
| 16 | Heart rate permutation entropy 5,2 | + |
| 17 | Lactate level | - |
| 18 | Oxygen Saturation mean db3 wavelet decomposition coefficient [a measure of the mean of the coefficients of a wavelet decomposition by Daubechies wavelet filter 3] | + |
| 19 | Heart rate POLVAR 5,3 | - |
| 20 | Heart rate POLVAR 5,4 | - |

GCS: Glasgow Coma Scale; mGCS: motor subscore of the Glasgow Coma Scale; eGCS: eye subscore of the Glasgow Coma Scale; SOFA: sequential organ failure assessment, HR: heart rate; (+): positive correlation; (-) negative correlation, POLVAR: measures the probability of obtaining a sequence of consecutive ones/zeros

**Figure 1. Study flow diagram.**

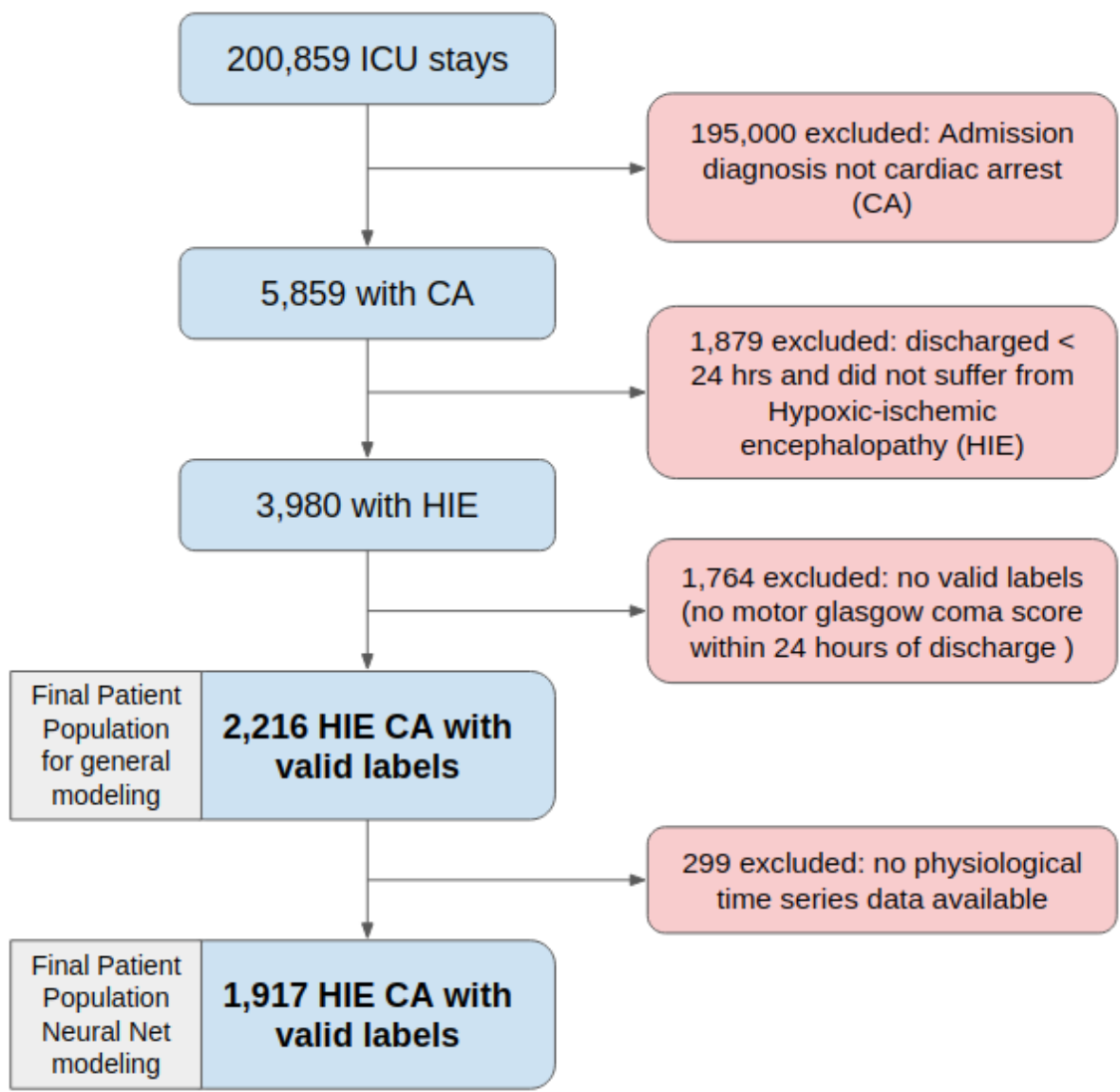

Selection process to achieve the final post-cardiac arrest population for machine learning from the eICU database.

**Fig. 2. Performance of seven machine learning algorithms used to predict post-CA neurological outcome.**

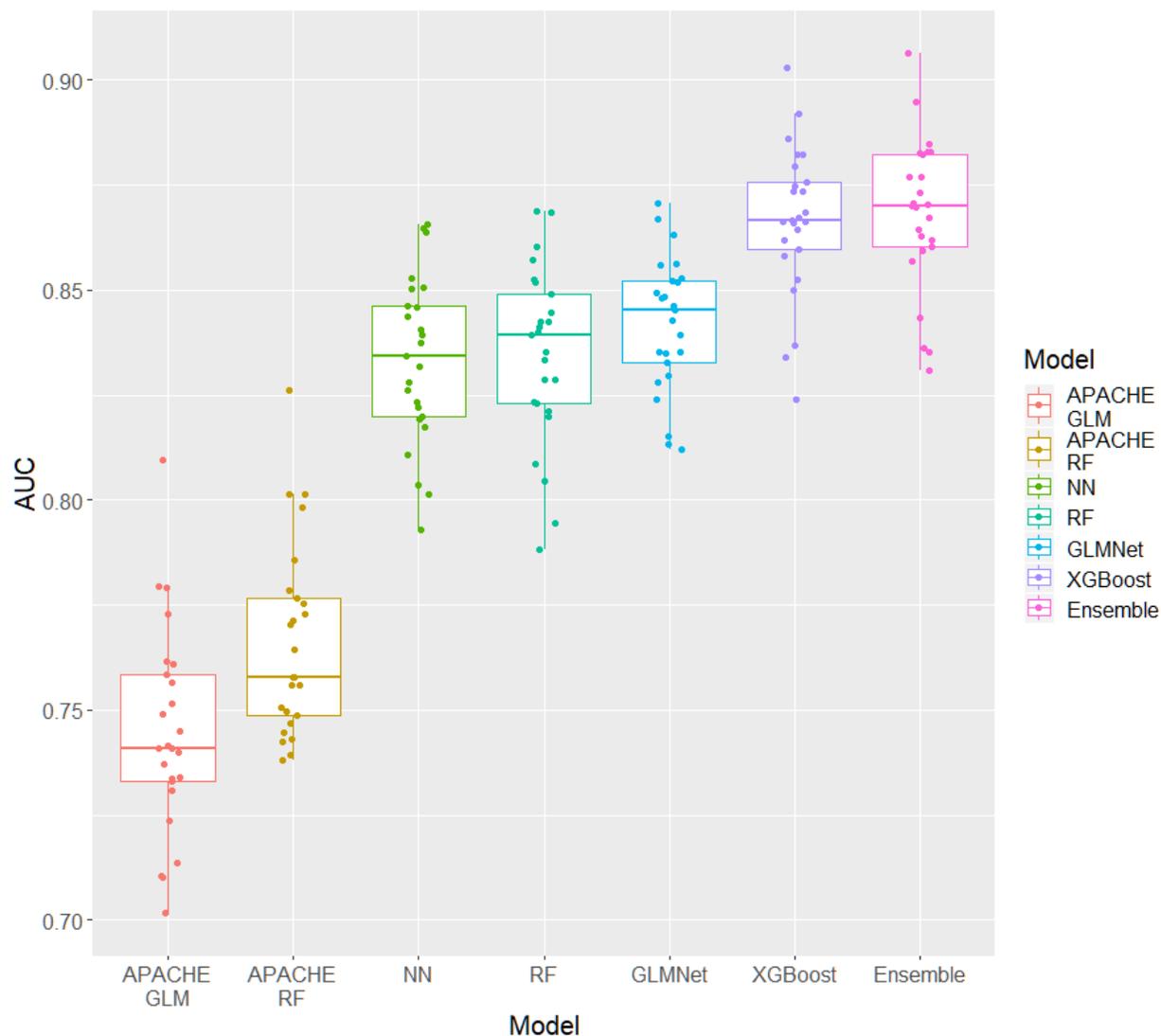

Each point represents the AUC of a different test of the nested resampling validation method. The corresponding AUC and statistical tests are in Table S1.

AUC: area under the receiver operating characteristic curve; GLM: generalized linear model, NN: neural network, RF: random forest; XGBoost: extreme gradient boosting; Ensemble: Ensemble model averaging the four best performing first-level models (XGBoost, GLM Elastic Net, RF, and NN).

**Fig. 3. Performance of seven machine learning algorithms used to predict post-CA survival**

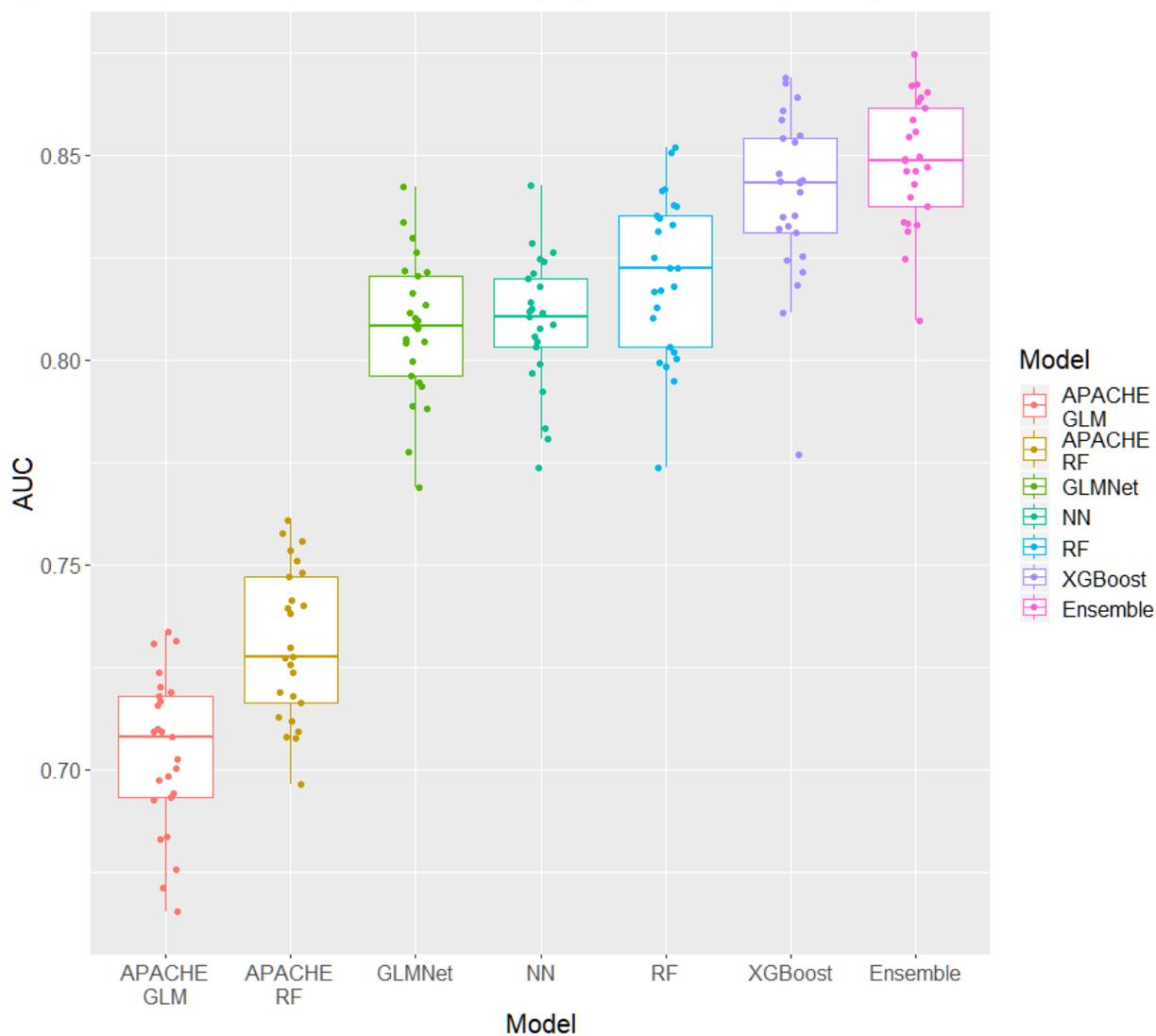

Each point represents the AUC of a different test of the nested resampling validation method. The corresponding AUC and statistical tests are in Table 4.

AUC: area under the receiver operating characteristic curve; APACHE: Acute Physiology and Chronic Health Evaluation; GLM: generalized linear model, NN: neural network, RF: random forest; XGBoost: extreme gradient boosting; Ensemble: Ensemble model averaging the four best performing first-level models (XGBoost, GLM Elastic Net, RF, and NN).

**Fig. 4. Best performing models for post-CA prediction**

A. Prediction of neurological outcome

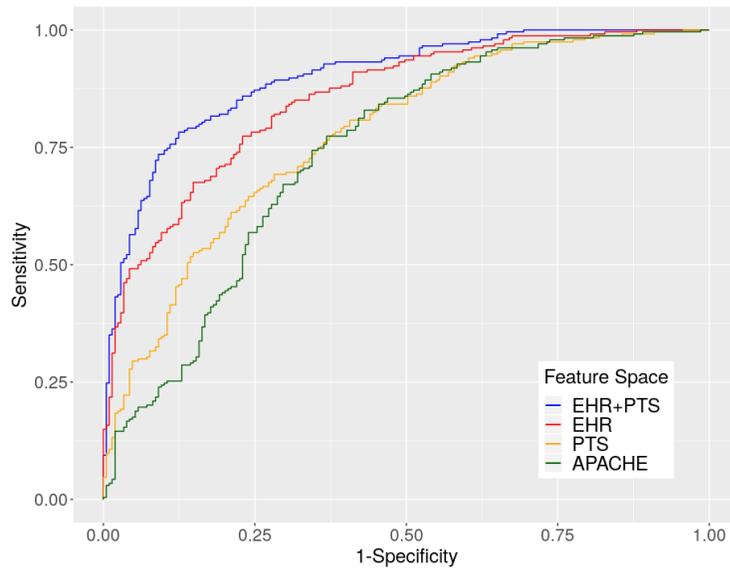

B. Prediction of survival

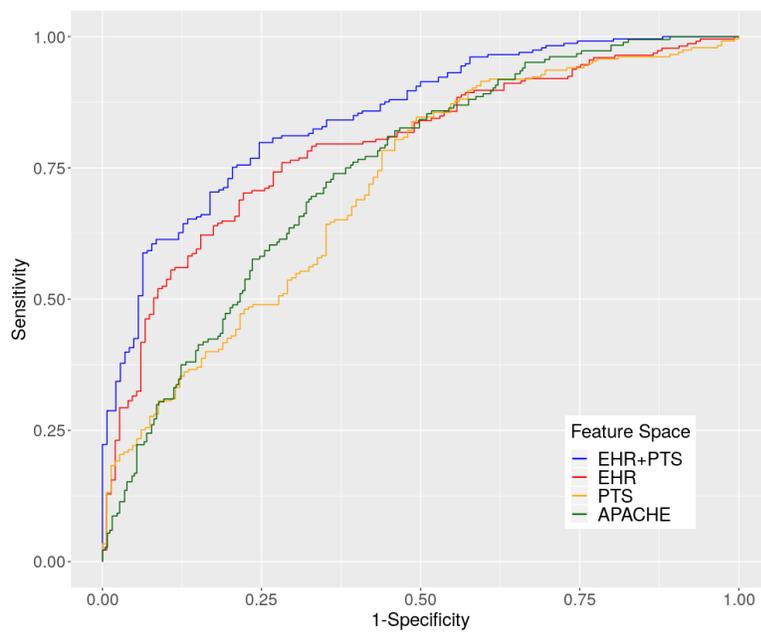

ROC of the best performing model (BPM) using different features: EHR only, PTS$_{24}$ only, both EHR and PTS, and the reference standard (APACHE) for neurological outcome prediction (A) and survival prediction (B).

APACHE: Acute Physiology and Chronic Health Evaluation; PTS: physiological time series, EHR: electronic health record, AUC: Area under the receiver operating characteristic curve

**Fig. 5. Relative importance of feature categories for prediction of neurological outcome**

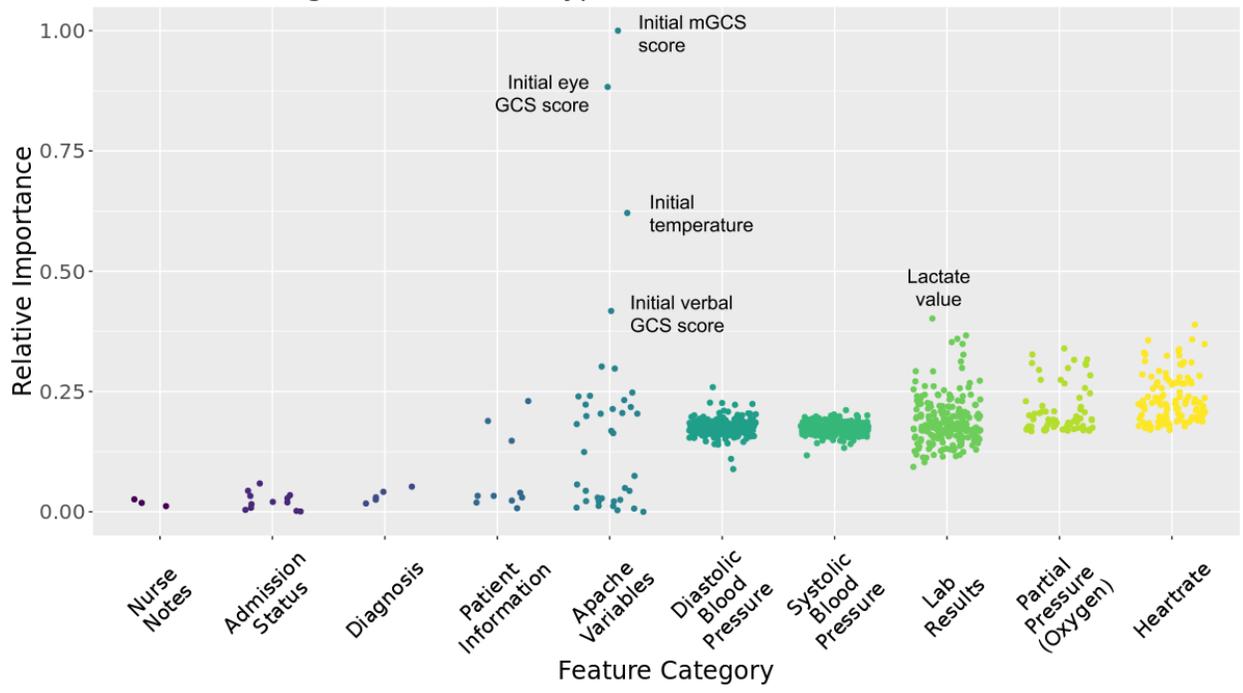

Each dot represents an individual feature used in the Ensemble model for neurological outcome prediction. Relative importance is based on the minimum depth of the maximum subtree using random forest.

GCS: Glasgow Coma Scale, mGCS: motor GCS subscore

**Fig. 6. Relative importance of physiological time series in the first 24 hrs after ICU admission**

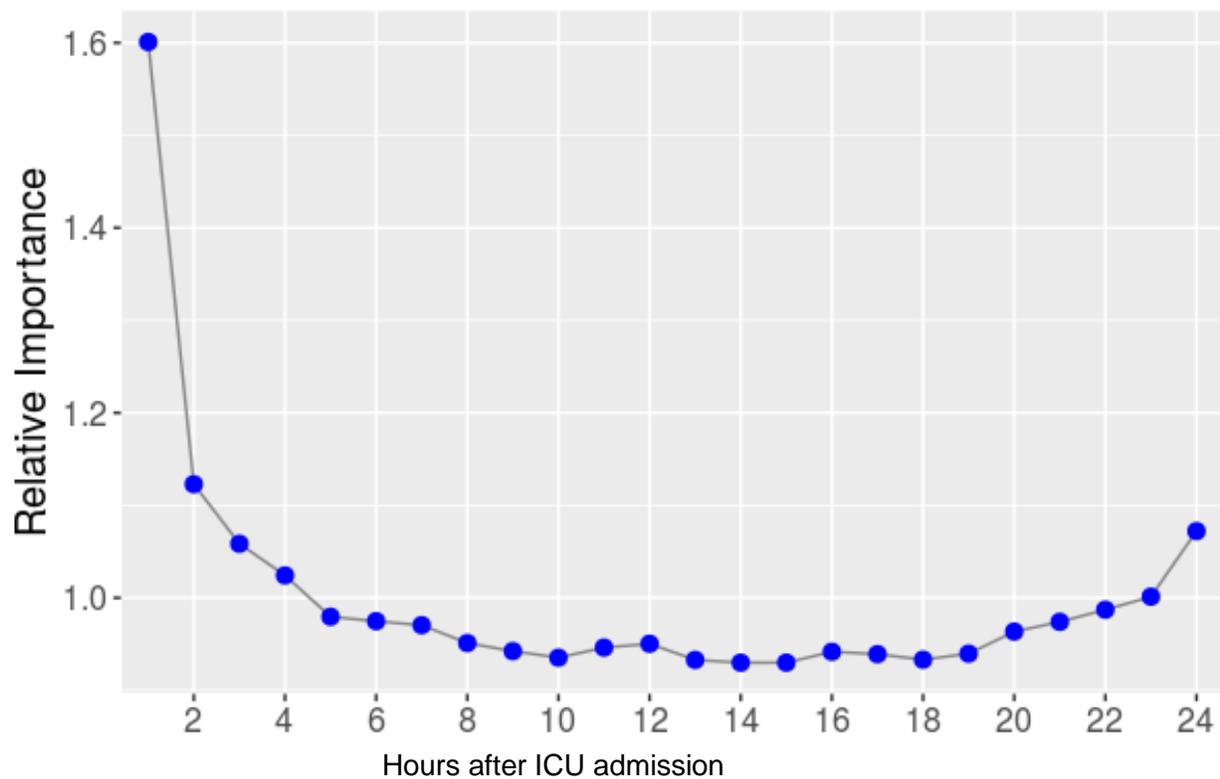

Relative predictive importance of PTS data collected in first 24 hours after ICU admission for prediction of neurological outcome in the ICU

**Fig. 7. Machine learning pipeline**

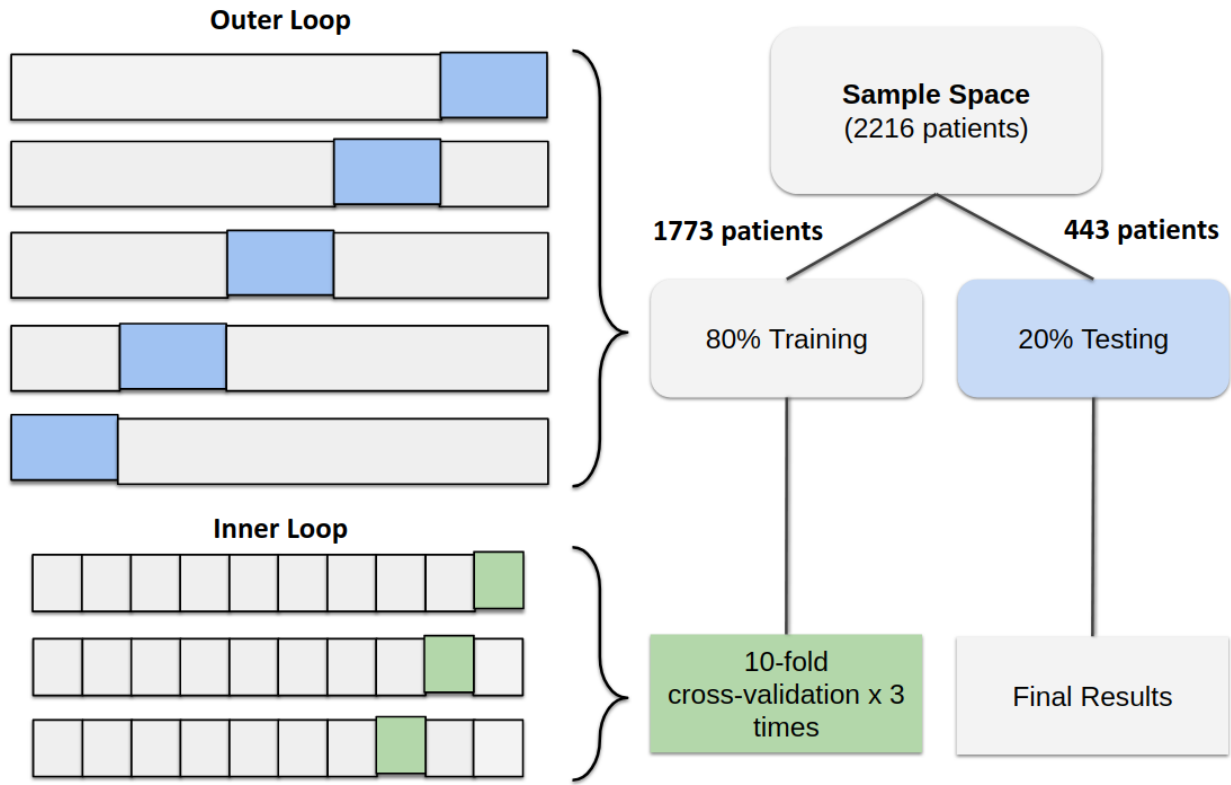

Nested cross validation in which the inner loop is used for hyperparameter tuning and the outer loop is to evaluate generalized model performance and compare among different models.

**Supplementary Materials**

**A Physiology-Driven Computational Model for Post-Cardiac Arrest Outcome Prediction**

Han Kim, Hieu Nguyen, Qingchu Jin, Sharmila Tamby, Tatiana Gelaf Romer, Eric Sung, Ran Li, Joseph Greenstein, Jose I. Suarez, Christian Storm, Raimond Winslow, Robert D. Stevens

**Contents**

Materials and Methods

Supplementary Tables and Figures:

## Materials and Methods

### *Evaluation Metrics*

Model performance was determined by evaluating sensitivity, specificity and discrimination (area under the receiver operating characteristic curve [AUC]). The AUC is a metric that is computed from sensitivity and specificity across different thresholds. In binary classification problems, the AUC falls between 0.5 and 1 where values closer to 1 represent higher performance of the model. Clinical models with an AUC of >0.8 or >0.9 are regarded as having good or excellent discrimination, respectively. For each model, we report the mean AUC along with 95% confidence intervals for the testing set. We also assessed the sensitivity and specificity for each model across multiple validation runs.

### *PTS outlier detection and imputation*

This process is illustrated in Fig. S1. We reasoned that outliers can be categorized as either an anomaly (artifact) or an accurate reading (real clinical events). To identify outliers, the sliding window median and median absolute deviation for PTS data were calculated. Within the window, any single point or continuous interval of points that falls outside of 3 absolute deviations from the sliding window median are considered potential outliers. Next, two rejecting bounds (lower and upper bounds deemed clinically implausible by physicians) are set and can be found in Table S3. For each potential outlier interval, the entire interval is removed if at least one point meets the above outlier criteria. The rationale behind these criteria is that outliers in the same interval are triggered by either a clinical event or a machine malfunction event. In any given temporal interval if it was shown that one outlier was artifact, then all other outliers in this outlier interval were considered artifact, and the whole interval was removed.

For PTS data imputation, the first step was to capture data in the nurse charting records. In many ICUs nurses manually record various features such as heart rate, blood pressure, and temperature and these data are then archived in the EHR. Inconsistencies can sometimes be observed between manually recorded data and PTS data; this may be due to offsets between the time vital signs are recorded and the physiological occurrence. However, when EHR and PTS data are strongly correlated, EHR data is a valuable resource for imputation of missing PTS data. We binned nurse charting data into 5 min intervals, mirroring the format of the PTS data. Next, the Pearson correlation for all overlapping time points between the EHR and PTS data was calculated. EHR data were used were used for imputation if the degree of correlation was >0.8 for >15 common timepoints. The remaining missing data were imputed using linear interpolation.

### *Neural network structures and training scheme*

Several types of neural network architectures including convolution neural net (CNN), recurrent neural net (LSTM, GRU) were tested on the PTS data. The results (not shown) demonstrated that CNN performed best. The CNN consisted of three one-dimensional convolutional layers with maximum pooling and ReLU activation layer. The kernel size for convolutional layer from first to third was 3, 5 and 5, respectively. One additional fully connected layer with sigmoid function was added after the last convolutional layer. The neural net selected to train on all data (EHR + HCTSA package derived features + PTS) was the combination of the same CNN for PTS and a fully connected (FC) layer (1014 to 1) for both EHR and HCTSA package derived features. The outputs of the convolutional net and FC layer are both one value. An additional FC layer (2 to 1) with a sigmoid function was added after those two layers in order to combine and give a final prediction. All steps were the same as with the other ML training approaches reported here, except that an additional subset was necessary for the neural net to define the optimal training epoch. We separated 12.5% of the training data as this additional subset.

### *Transfer learning*

The pre-train classification task is the mortality prediction of the entire available population. The available number of patient observations decreased from 200,859 to 140,200 due to inconsistent availability of EHR and PTS. Because of the extremely high computational cost to obtain the HCSTA package derived features, it was not included in the pre-train task. The data is split into a training set (85%), validation set (7.5%), and test set (7.5%). The same preprocessing steps were performed as described above. A CNN was selected for PTS data (Fig. S2B). A three-layer FC neural net was selected for EHR data (Fig. S2A). A combined (FC + convolutional) net was selected for the combined data (Fig. S2C). Due to the severe outcome class imbalance in this dataset (Alive: Expired ≈ 10:1), the weighted focal loss method was used. Thirty epochs were used for each training. The optimal epoch was decided with maximum validation set AUC. The pre-trained convolutional model for PTS was then further trained on our focused patient population (1,917 patients) for two labels: neurological outcome label and mortality label. We used the pre-trained model for both EHR+PTS for transfer learning because it does not contain the HCTSA package-derived features. We initialized a portion of weights of the model for PTS+EHR from CNN model for PTS alone, and then trained the model (Fig. S2D).

### *Recurrent neural net with attention layer*

Attention layer has been used on healthcare image data for lesion detection *(52)*. Shickel et. al. utilized self-attention to gain knowledge on the relative importance for each hour of the PTS data *(53)*. We implemented this approach in our model with a different neural net architecture, to attain the relative predictive importance of PTS information from each hour in Fig. 6. The model architecture, shown in Fig. S2E consists of 6 parallel GRU units with an attention layer (α) on top of each hidden layer. The input is PTS data alone

## Supplementary Tables and Figures

| Table S1. Statistical comparison of the discrimination of the seven machine learning algorithms used to predict post-CA neurological outcome. | | |
|---|---|---|
| **Feature Space** | **Mean AUC** | **Wilcoxon Ranked Sum p-value*** |
| **APACHE GLM** | 0.745 | |
| | | 0.00265000 |
| **APACHE RF** | 0.766 | |
| | | 0.00000006 |
| **PTS+EHR NN** | 0.833 | |
| | | 0.68600000 |
| **PTS+EHR RF** | 0.835 | |
| | | 0.22400000 |
| **PTS+EHR GLMnet** | 0.842 | |
| | | 0.00000487 |
| **PTS+EHR XGBoost** | 0.867 | |
| | | 0.65810000 |
| **PTS+EHR Ensemble** | 0.868 | |

* p-value evaluated are given for each pair of adjacent AUCs

AUC: area under the receiver operating characteristic curve; APACHE: Acute Physiology and Chronic Health Evaluation; EHR: electronic health record; PTS: physiologic time series; GLM: generalized linear model, NN: neural network, RF: random forest; XGBoost: extreme gradient boosting; Ensemble: Ensemble model averaging the four best performing first-level models (XGBoost, GLM Elastic Net, RF, and NN). See also Figure 2.

| Feature Space | Mean AUC | Wilcoxon Ranked Sum p-value* |
|---|---|---|
| **Table S2.** Statistical comparison of the performance of the seven machine learning algorithms used to predict post-CA survival. | | |
| **APACHE GLM** | 0.704 | |
| | | 0.00003390000000 |
| **APACHE RF** | 0.731 | |
| | | 0.00000000000002 |
| **PTS+EHR GLMnet** | 0.808 | |
| | | 0.74400000000000 |
| **PTS+EHR NN** | 0.809 | |
| | | 0.03900000000000 |
| **PTS+EHR RF** | 0.821 | |
| | | 0.00075900000000 |
| **PTS+EHR XGBoost** | 0.840 | |
| | | 0.10350000000000 |
| **PTS+EHR Ensemble** | 0.849 | |

\* p-value evaluated are given for each pair of adjacent AUCs

AUC: area under the receiver operating characteristic curve; APACHE: Acute Physiology and Chronic Health Evaluation; EHR: electronic health record; PTS: physiologic time series; GLM: generalized linear model, NN: neural network, RF: random forest; XGBoost: extreme gradient boosting; Ensemble: Ensemble model averaging the four best performing first-level models (XGBoost, GLM Elastic Net, RF, and NN). See also Fig. 3

**Table S3. Statistical comparison of model performance using different feature sets**

| (A) Neurological Outcome | | |
|---|---|---|
| **Feature Space** | **Mean AUC** | **Wilcoxon Ranked Sum p-value** |
| APACHE GLM | 0.745 | |
| | | 0.0000674000 |
| PTS | 0.775 | |
| | | 0.0000000002 |
| EHR | 0.831 | |
| | | 0.0000000271 |
| EHR + PTS | 0.868 | |
| (B) Survival | | |
| PTS | 0.675 | |
| | | 0.0000133000 |
| APACHE GLM | 0.704 | |
| | | 0.0000000000 |
| EHR | 0.799 | |
| | | 0.0000000001 |
| EHR + PTS | 0.848 | |

AUC: area under the receiver operating characteristic curve; APACHE: Acute Physiology and Chronic Health Evaluation; EHR: electronic health record; PTS: physiologic time series; GLM: generalized linear model, NN: neural network, RF: random forest; XGBoost: extreme gradient boosting; Ensemble: Ensemble model averaging the four best performing first-level models (XGBoost, GLM Elastic Net, RF, and NN). See also Fig. 4

**Table S4. Clinically determined lower and upper bounds for physiological time series**

| | HR (beat/min) | RR (breaths/min) | Diastolic BP (mmHg) | Systolic BP (mmHg) | SpO2 |
|---|---|---|---|---|---|
| Lower bound | 30 | 6 | 20 | 50 | 60 |
| Upper bound | 200 | 50 | 100 | 220 | 100 |

PTS: physiological time series; HR: heart rate; RR: respiratory rate; BP: blood pressure; SpO2 pulse oximetry saturation

**Table S5. Transfer learning pre-trained model results (Survival prediction)**

| Data type | Neural Network type | Validation set AUC | Test set AUC |
|:---:|:---:|:---:|:---:|
| **EHR** | Fully connected neural net | 0.89 | 0.87 |
| **PTS** | Convolutional neural net | 0.84 | 0.85 |
| **EHR+PTS** | Fully connected + Convolutional neural net | 0.90 | 0.90 |

EHR: electronic health record, PTS: physiological time series, AUC: area under the curve

**Fig. S1. Physiological time series denoising and imputation**

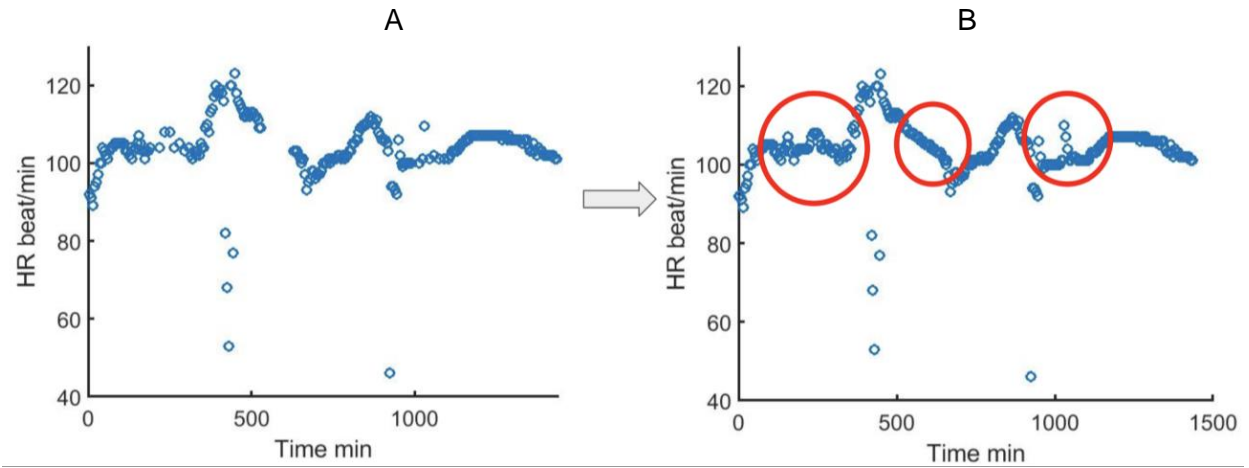

An example of the heart rate physiological time series (PTS) denoising and imputation. Panel A is one patient's PTS original heart rate data with missing values and outliers. Panel B shows the results after denoising and imputation.

**Fig. S2. Architecture of neural networks used in this study**

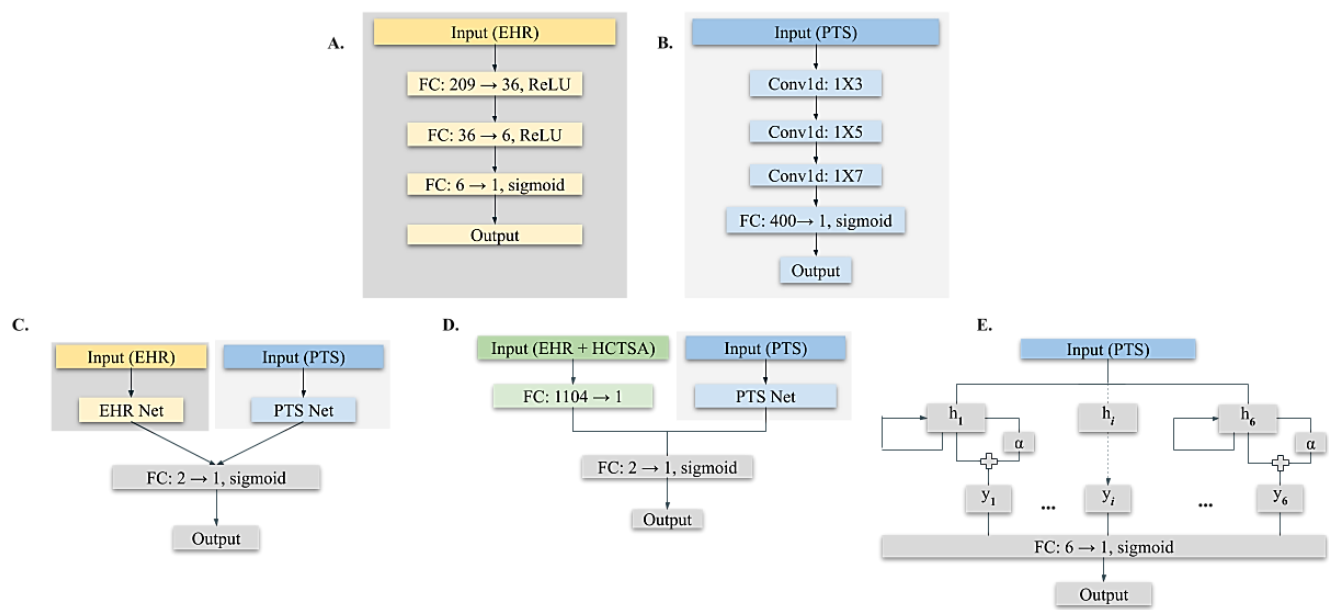

A and B represent the structure of neural networks to process EHR and PTS data, respectively. C is the architecture for pre-training for the source population (140,200 non-CA patients), and D is the architecture for training on the target population (1917 CA-patients). E is the architecture for the GRU with attention layer (represented as α) which was used to obtain the relative predictive importance of each hour within the first 24 hours.

EHR: electronic health record, PTS: physiologic time series, FC: fully connected layer, Conv: convolution layer, GRU: Gated recurrent unit

**Fig. S3. Discrimination of transfer learning models**

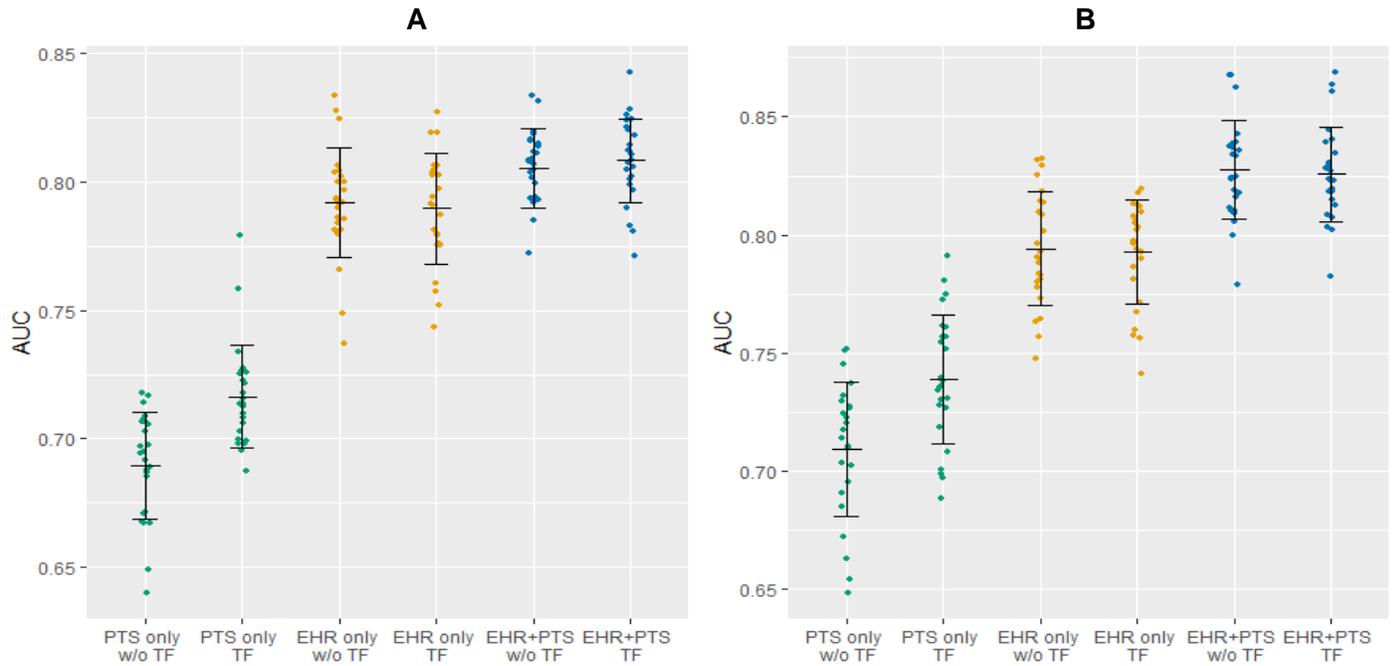

Transfer learning results for survival prediction (A) and neurological outcome prediction (B). Green points: PTS data, yellow points: EHR data, blue points: EHR data + PTS data + HCTSA derived features. TF: transfer learning; EHR: electronic health record, PTS: physiological time series, AUC: area under the curve, TF: transfer learning

**Fig. S4. Relative importance of feature categories for survival prediction.**

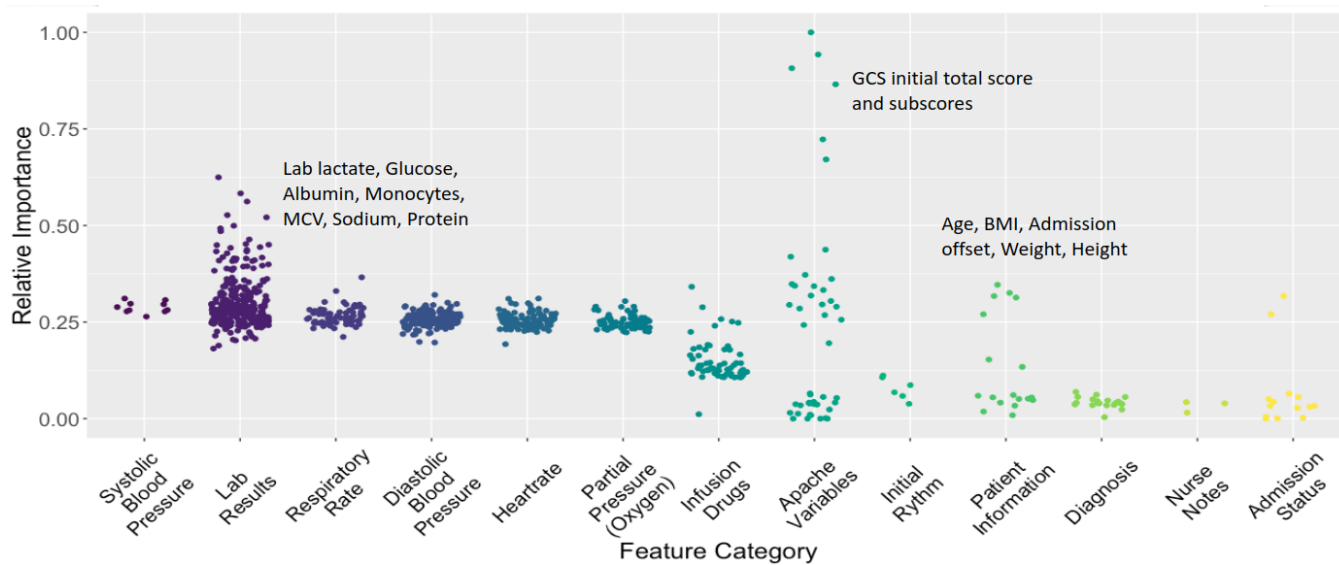

Relative importance of feature categories from the survival prediction ensemble model. Each dot represents an individual feature. Relative importance is based on the minimum depth of the maximum subtree found in random forest.